\documentclass{article}

\usepackage[preprint]{neurips_2023}
\usepackage[utf8]{inputenc}
\usepackage[T1]{fontenc}
\usepackage[hyphens]{url}
\usepackage{booktabs}
\usepackage{amsfonts}
\usepackage{amsmath}
\usepackage{amssymb}
\usepackage{nicefrac}
\usepackage{microtype}
\usepackage{xcolor}
\usepackage{multirow}
\usepackage[colorlinks=true,
            linkcolor=blue,
            citecolor=blue,
            urlcolor=blue]{hyperref}
\usepackage{graphicx}
\usepackage{xspace}
\usepackage{enumitem}
\usepackage{tikz}
\usetikzlibrary{arrows.meta,positioning,decorations.pathreplacing}
\definecolor{h2oyellow}{HTML}{F5C518}
\definecolor{trained}{HTML}{C1440E}
\definecolor{qblue}{HTML}{E4F1FC}
\definecolor{qedge}{HTML}{9EC3E6}
\usepackage{pgfplots}
\pgfplotsset{compat=1.18}
\usepackage{placeins}
\usepackage{listings}
\graphicspath{{figures/}}

\newcommand{\tabh}{\mbox{\textit{TabH2O}}\xspace}
\newcommand{\tlm}{\mbox{\textsc{TLM}}\xspace}
\newcommand{\Rb}{\mathbb{R}}
\newcommand{\Dtrain}{\mathcal{D}_{\mathrm{train}}}

\title{Emergent Abilities of Tabular Foundation Models:\\From Tables, to Text, Images, and Robotics}

\author{%
  \textbf{Marcos V. Conde}$\quad$\textbf{Laura Fink}$\quad$\textbf{Olivier Grellier}$\quad$\textbf{Sri Ambati}\\%
  H2O.ai\\%
  \texttt{\{firstname.lastname, sri\}@h2o.ai}\\%
}

\begin{document}

\maketitle

\begin{abstract}
Tabular foundation models (TFMs) now rival or outperform tuned gradient-boosted trees on standard tabular benchmarks, performing in-context inference in one forward pass after pretraining on millions of synthetic tables. Yet nearly all TFM research remains confined to tables. We explore new capabilities and applications of TFMs across \emph{four domains absent from pretraining}: natural language, images, motion forecasting, and sensor-driven control. Because a deployed TFM standardises each column and averages over column permutations, it operates on numerical row geometry with no trained multimodal projector. Writing sentence embeddings, flattened pixels, motion histories, or LiDAR scans as columns lets one frozen checkpoint (\tabh) reach $93.3\%$ on IMDb and $100$-shot MNIST, add $24$ balanced-accuracy points on mixed text-and-tabular fraud detection, beat matched linear, tree, and neural baselines on motion forecasting and robot cloning, and steer a car from LiDAR. We also measure where transfer breaks down: scarce context, near-linear targets, and distribution shift.
\end{abstract}

\section{Introduction}
\label{sec:intro}

Tabular Foundation Models (TFMs) have replaced gradient-boosted trees~\cite{chen2016xgboost,ke2017lightgbm,prokhorenkova2018catboost}---the decade-long default across credit, fraud, clinical risk, and demand forecasting---as the strongest predictors on standard tabular benchmarks. TabPFN~\cite{hollmann2023tabpfn,hollmann2025tabpfnv2}, TabICL~\cite{qu2025tabicl,qu2026tabiclv2}, TabDPT~\cite{ma2025tabdpt}, \tabh~\cite{pfeiffer2026tabh2o}, TabFM~\cite{google2026tabfm}, EXAONE Tabular~\cite{eo2026exaone}, and Causilo~\cite{numsai2026causilo} implement prior-data fitted networks~\cite{muller2021pfn} trained to approximate posterior predictive inference by minimising
\begin{equation}
\mathcal{L}(\theta) \;=\; \mathbb{E}_{\mathcal{D}\sim p(\cdot)}\!\left[-\log q_\theta\!\left(y_{\mathrm{test}} \mid x_{\mathrm{test}}, \Dtrain\right)\right],
\label{eq:pfn}
\end{equation}
where $\mathcal{L}(\theta)$ is the expected negative log-likelihood, $q_\theta$ is the TFM predictive distribution with parameters $\theta$, $\Dtrain$ is the labelled context table, $(x_{\mathrm{test}}, y_{\mathrm{test}})$ is a held-out query pair from $\mathcal{D}$, and $p(\cdot)$ is a prior over millions of \emph{synthetic} datasets from random structural causal models~\cite{pearl2009causality}. At inference, a TFM takes a labelled table and query rows and returns calibrated predictions in one forward pass, outranking tuned ensembles and AutoML pipelines on TabArena~\cite{erickson2025tabarena}.

Yet almost all TFM work stays inside the tabular box: training on tables and benchmarking on tables. Recent efforts scale row and column counts or broaden tabular suites, while free text is dropped or hashed because TFMs cannot read column names---a limitation TabICL\,v2 lists first~\cite{qu2026tabiclv2}---and images, trajectories, and sensor streams are left out entirely. This narrow scope leaves a central question open:
\begin{center}
\setlength{\fboxsep}{7pt}%
\fcolorbox{qedge}{qblue}{%
\parbox{0.94\linewidth}{\centering\emph{Does pretraining on synthetic causal tables unlock general relational inference over numerical data, enabling new applications far beyond conventional tables?}}}
\end{center}

We explore that question across \textbf{four domains absent from pretraining}: \emph{language} (text and mixed tables), \emph{vision} (flattened pixels and visual embeddings), \emph{motion} (handwriting, pedestrian trajectories, and robot cloning), and \emph{control} (LiDAR steering). If we can project the information (text, images, motion, etc.) into a structured $\Rb^{d}$ embedding, TFMs can work on it (Table~\ref{tab:overview}; Appendix~\ref{app:invariances}). TFMs are row-invariant, but not column-invariant: the order of the features matters.

\begin{table}[t]
\centering
\caption{\textbf{One frozen TFM, four domains, no gradient steps.} Every task is turned into ordinary rows and columns and given to the same \tabh{} checkpoint by in-context learning. The control is the best simple model on the \emph{identical} columns (linear probe, ridge, or the same table without text).}
\label{tab:overview}
\small
\setlength{\tabcolsep}{4pt}
\begin{tabular}{llllcc}
\toprule
Domain & Task & One row is\dots & Metric & Control & \tabh \\
\midrule
Language & IMDb sentiment & review embedding, $96$ cols & acc.\ (\%) & 93.1 & \textbf{93.3} \\
 & Fake job ads & metadata $+$ ad text & bal.\ acc.\ (\%) & 54.6 & \textbf{78.6} \\
 & LIAR truthfulness & metadata $+$ statement text & acc.\ (\%) & 25.0 & \textbf{31.4} \\
Vision & MNIST, $100$/class & $22{\times}22$ pixels, $484$ cols & acc.\ (\%) & 85.1 & \textbf{93.3} \\
Motion & LASA handwriting & $16$-step history, $64$ cols & ADE $\downarrow$ & 0.423 & \textbf{0.403} \\
 & PushT robot cloning & state $+$ keypoints, $18$ cols & action $L_2\downarrow$ & 21.2 & \textbf{13.6} \\
Control & Cave driving & $180$ LiDAR ranges & acc.\ (\%) & 82.8 & \textbf{90.8} \\
\bottomrule
\end{tabular}
\end{table}

\paragraph{Emergent abilities and their limits.} We use \emph{emergent} in a descriptive sense: capabilities that appear at inference despite being absent from the synthetic prior on table pre-training, which contains no text, images, trajectories, or control loops. Yet the same frozen checkpoint classifies reviews, reads digits without convolutions, forecasts curved motion, clones robotic actions, and steers from LiDAR (Table~\ref{tab:overview}). Crucially, we also characterise the exact boundaries of this transfer: in-context inference requires enough support rows to identify the mapping, a genuinely nonlinear target structure, and approximate exchangeability between context and query rows.

\paragraph{Relation to concurrent work.} Nakerst et al.~\cite{nakerst2026nontabular} showed that TabPFN\,v3 classifies flattened MNIST and identifies languages, and a follow-up~\cite{nakerst2026cave} steered a simulated car from LiDAR. We reproduce both phenomena on a different TFM, evaluate them against matched neural and tree baselines on identical features, and extend the study to real-world mixed text-plus-tabular tables, trajectory forecasting, and robotic behaviour cloning.

\paragraph{Experimental protocol.} All experiments use a single frozen \tabh~\cite{pfeiffer2026tabh2o} checkpoint ($29.2$M parameters), evaluated within our current GPU memory budget of up to $500$ columns and $100{,}000$ rows. Every benchmark compares \tabh{} against matched baselines on identical columns using official test splits wherever available. Appendix~\ref{sec:properties} records the invariances and the recipe. Ablations and dataset splits are in the supplementary material.

\section{TFMs for language: Tabular Language Models}
\label{sec:language}

\begin{figure}[t]
\centering
\resizebox{\textwidth}{!}{%
\begin{tikzpicture}[
  font=\sffamily\small,
  box/.style={draw=black, line width=0.9pt, rounded corners=4pt, minimum height=1.2cm,
              align=center, fill=white, inner xsep=7pt},
  frozen/.style={box, fill=h2oyellow!40},
  dark/.style={box, fill=black, text=white},
  flow/.style={-{Stealth[length=6pt,width=5pt]}, line width=1.1pt},
  note/.style={font=\sffamily\scriptsize, text=gray!55!black, align=center},
  head/.style={font=\sffamily\Large, text=black, align=left},
  cell/.style={draw=gray!55, minimum width=0.30cm, minimum height=0.26cm, inner sep=0pt, line width=0.3pt},
]

\node[head, anchor=west] at (-0.4, 3.15) {\textbf{(a)}\; the Tabular-Language Model pipeline};

\node[note, anchor=south] at (0.95, 2.15) {raw table};
\foreach \r in {0,1,2,3} {
  \pgfmathsetmacro{\yy}{1.75-0.3*\r}
  \node[cell, fill=gray!12] at (0.30,\yy) {};
  \node[cell, fill=gray!12] at (0.62,\yy) {};
  \node[cell, fill=h2oyellow!55, minimum width=0.62cm] at (1.11,\yy) {};
  \node[cell, fill=gray!30] at (1.60,\yy) {};
}
\node[note, anchor=north] at (0.95, 0.58) {$p$ numeric\;\textbar\;text\;\textbar\;$y$};

\draw[flow] (1.95, 1.60) -- (12.00, 1.60);
\node[note, anchor=south] at (6.00, 1.70) {structured stream $x\in\Rb^{p}$ (untouched)};

\draw[flow] (1.95, 1.00) -- (2.30, 1.00) -- (2.30, -0.55) -- (2.80, -0.55);
\node[frozen, minimum width=3.8cm] (enc) at (4.70,-0.55)
  {Text Encoder\\(frozen)\\[-1pt]\scriptsize $\varepsilon$, BGE-small, 33M, 384-d};
\draw[flow] (6.60,-0.55) -- (7.40,-0.55);
\node[box, minimum width=3.8cm] (pca) at (9.30,-0.55)
  {projection $\pi_k$\\[-1pt]\scriptsize PCA, train rows, \textbf{no labels}};
\node[note, anchor=north] at (4.70,-1.35) {no gradients};
\node[note, anchor=north] at (9.30,-1.35) {sets the interface width $k$};

\draw[flow] (11.20,-0.55) -- (12.00,-0.55) -- (12.00, 0.34);
\draw[flow] (12.00, 1.60) -- (12.00, 0.76);
\node[circle, draw=black, fill=white, text=black, line width=0.9pt, inner sep=1.5pt, font=\sffamily\footnotesize\bfseries] (m) at (12.00,0.55) {$+$};
\draw[flow] (m.east) -- (14.55,0.55);
\node[note, anchor=south] at (13.40,0.70) {$Z\in\Rb^{n\times(p+k)}$};

\node[box, text=black, minimum width=2.9cm, minimum height=2.0cm] (tfm) at (16.10,0.55)
  {\textbf{\large TFM}\\[4pt] in-context learning};
\draw[flow] (17.55,0.55) -- (18.40,0.55);
\node[frozen, minimum width=1.7cm, minimum height=1.4cm] (out) at (19.30,0.55) {$\hat{y}$\\[-1pt]\scriptsize $+$ quantiles};
\node[note, anchor=north, text width=3.6cm] at (16.10,-0.65)
  {labels enter \emph{only} as in-context rows};

\begin{scope}[yshift=-3.9cm]
  \node[head, anchor=west] at (-0.4, 1.55) {\textbf{(b)}\; where the training happens};
  \node[note, anchor=east, text=black] at (2.20, 0.50) {\textbf{VLM}};
  \node[frozen, minimum height=0.9cm, minimum width=3.2cm, align=center] (v1) at (4.00,0.50)
    {\footnotesize Image Encoder\\CLIP (frozen)};
  \node[box, fill=trained!15, draw=trained, minimum height=0.9cm, minimum width=4.0cm] (v2) at (8.40,0.50)
    {\footnotesize\textcolor{trained}{\textbf{Learn Projection}}};
  \node[box, minimum height=0.9cm, minimum width=3.2cm] (v3) at (13.20,0.50) {\footnotesize LLM};
  \draw[flow] (v1) -- (v2); \draw[flow] (v2) -- (v3);
  \node[note, anchor=west, text width=4.2cm] at (15.20,0.50)
    {alignment to a \emph{learned}\\token-embedding manifold};
  \node[note, anchor=east, text=black] at (2.20,-0.90) {\textbf{TLM}};
  \node[frozen, minimum height=0.9cm, minimum width=2.8cm, align=center] (t1) at (4.00,-0.90)
    {\footnotesize Text Encoder\\(frozen)};
  \node[box, fill=trained!15, draw=trained, minimum height=0.9cm, minimum width=4.0cm] (t2) at (8.40,-0.90)
    {\footnotesize\textcolor{trained}{\textbf{Fit PCA}}};
  \node[box, minimum height=0.9cm, minimum width=3.2cm] (t3) at (13.20,-0.90) {\footnotesize TFM};
  \draw[flow] (t1) -- (t2); \draw[flow] (t2) -- (t3);
  \node[note, anchor=west, text width=4.2cm] at (15.20,-0.90)
    {no alignment needed:\\the input space is $\Rb^{k}$};
\end{scope}
\end{tikzpicture}}
\caption{\textbf{A \tlm{} is a VLM with the projector deleted.} \textbf{(a)}~Structured columns pass through untouched; free text is encoded by a frozen sentence encoder and projected to $k$ dimensions by a PCA fitted on training rows without labels. The concatenation is an ordinary numeric table, which a frozen TFM consumes by in-context learning. \textbf{(b)}~A vision-language model must \emph{train} a projector because an LLM reads a learned token-embedding manifold. A TFM reads raw $\Rb^{k}$, so the same slot is filled by an unsupervised projection (Section~\ref{sec:language}).}
\label{fig:arch}
\end{figure}
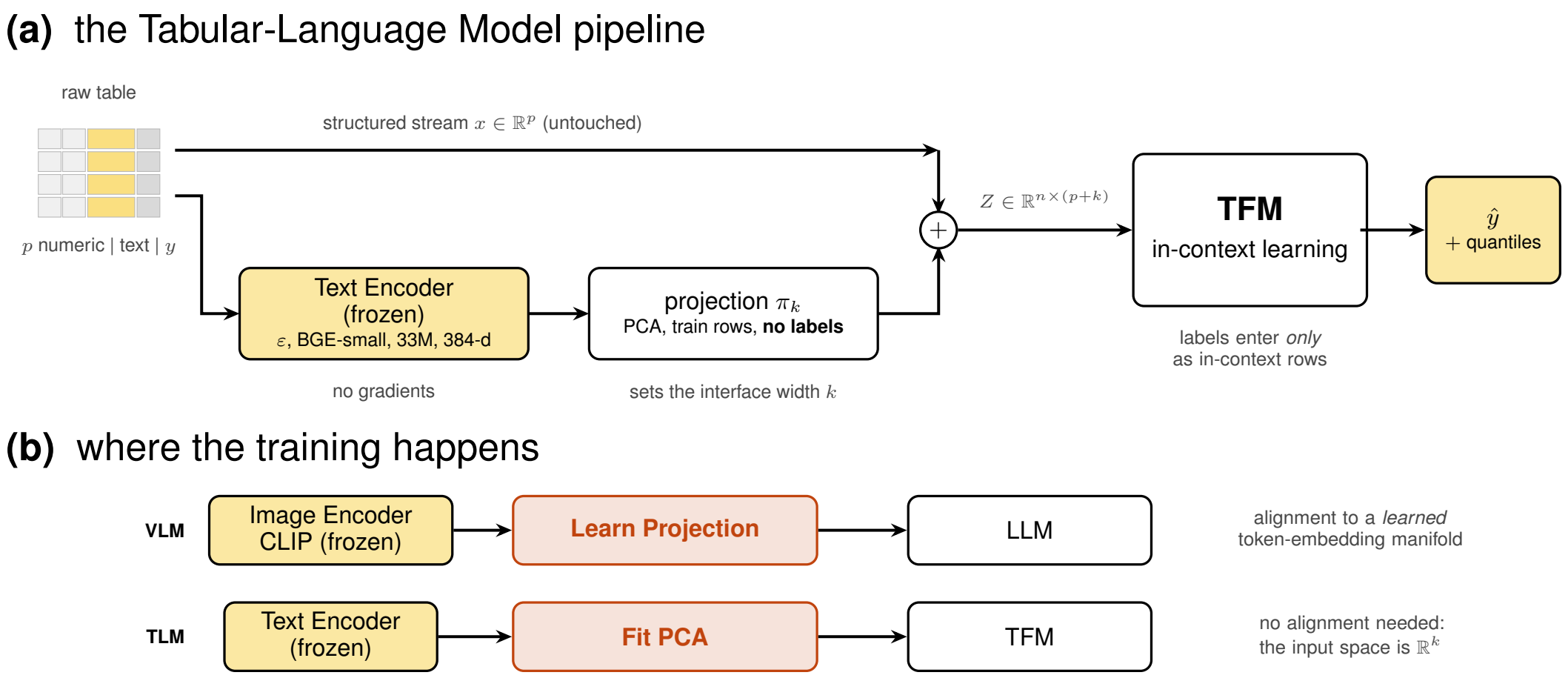

Vision-language models (VLMs) connect a frozen image encoder to a frozen language model through a \emph{trained} projector~\cite{liu2023llava,alayrac2022flamingo} because a language model reads a learned token-embedding manifold where an arbitrary feature vector must first be aligned. A TFM, by contrast, has no learned token manifold: it takes a plain $n\times m$ matrix, $z$-scores each column independently, and averages over column permutations, so a block of sentence embeddings already lies in its native input space.

That observation gives the \textbf{Tabular Language Model} (\tlm) of Figure~\ref{fig:arch}. Each text cell is encoded by a frozen sentence encoder (BGE-small~\cite{xiao2024bge}, $384$ dimensions), a PCA fitted on the training rows \emph{without labels} reduces it to $k{=}96$ columns, and the result is concatenated with whatever structured columns the table already has. A frozen TFM then predicts by in-context learning. Nothing is trained by gradient descent, and there is no projector: the PCA is the only object fitted to the dataset. We choose $k$ by the width rule of Appendix~\ref{app:invariances}; the full sweep is in the supplementary material.

\paragraph{Text-only classification.} We first evaluate \tabh{} on pure text classification: UCI Sentiment~\cite{kotzias2015sentiment} (short product, film and restaurant reviews), IMDb~\cite{maas2011imdb} (long film reviews), three TweetEval tasks~\cite{barbieri2020tweeteval} (sentiment, hate speech, offensive language), and GoEmotions~\cite{demszky2020goemotions} (Reddit comments, $28$ emotions). Every predictor sees the same frozen features. A \emph{linear probe} (logistic regression) on the same $96$ columns isolates what the TFM adds; a zero-shot \emph{label prompt} (cosine similarity to encoded class names) shows what the encoder knows on its own.

\begin{table}[t]
\centering
\caption{\textbf{Text-only classification with a frozen encoder and a frozen TFM.} The linear probe and \tabh{} receive the identical $96$-column PCA table. Official test splits except UCI, which has none (sealed $597$-row holdout). Best in bold.}
\label{tab:classification}
\small
\setlength{\tabcolsep}{4.5pt}
\begin{tabular}{llrccc}
\toprule
Dataset & Metric & Train\,/\,test & Label prompt & Linear probe & \tabh \\
\midrule
UCI Sentiment & accuracy & 2.4k\,/\,0.6k & 94.81 & 95.64 & \textbf{96.31} \\
IMDb & accuracy & 25k\,/\,25k & 88.69 & 93.07 & \textbf{93.32} \\
TweetEval sentiment & macro recall & 47.6k\,/\,12.3k & 59.43 & 67.49 & \textbf{68.95} \\
TweetEval hate & macro F1 & 10k\,/\,3.0k & 49.92 & \textbf{57.01} & 53.18 \\
TweetEval offensive & macro F1 & 13.2k\,/\,0.9k & 49.01 & \textbf{73.99} & 72.93 \\
\bottomrule
\end{tabular}
\end{table}

Table~\ref{tab:classification} shows that a model pretrained only on synthetic tables classifies real sentiment: $96.3\%$ on UCI and $93.3\%$ on IMDb, above the $88.9\%$ of the original IMDb paper~\cite{maas2011imdb} and ahead of the linear probe on the same columns. Frozen \tabh{} leads whenever context is sufficient and train and test splits are approximately exchangeable. It loses on TweetEval hate, whose test set was collected separately from its training set: the TFM assumes context and query are exchangeable, and under this covariate shift it over-predicts \texttt{hate}. Paired McNemar tests (supplementary material) confirm this is the sole significant loss among eight comparisons; the offensive-language gap is a tie.

\begin{table}[t]
\centering
\caption{\textbf{Fine-grained classification (GoEmotions) and mixed-table regression (Mercari 100K).} \textbf{Left:} GoEmotions $28$-way accuracy on the official test split ($5{,}427$ comments; $48{,}836$ training); ``surface'' sees five length and punctuation counts, and the linear column is a probe. \textbf{Right:} Mercari 100K log-price $R^2$ ($\log(1{+}\mathrm{price})$) on the official $25{,}000$-item test split ($20$k stratified training); the linear column is ridge. Frozen \tabh{} leads on every text and fused pack.}
\label{tab:goemotions-main}
\footnotesize
\setlength{\tabcolsep}{2.5pt}
\begin{minipage}[t]{0.47\linewidth}
\vspace{0pt}
\centering
\begin{tabular}{@{}lrrr@{}}
\toprule
\multicolumn{4}{@{}l}{\textup{\textbf{GoEmotions}}} \\
\multicolumn{4}{@{}l}{\emph{acc.\ (\%)\,$\uparrow$ \ ($5{,}427$ test comments)}} \\
\midrule
Columns & Cols & Linear & \tabh \\
\midrule
Surface counts only & 5 & 33.20 & 33.28 \\
Text (BGE$\to$PCA-96) & 96 & 47.13 & \textbf{48.11} \\
Fused: metadata $+$ text$^{\dagger}$ & 101 & 51.00 & \textbf{51.89} \\
\bottomrule
\end{tabular}
\end{minipage}\hfill
\begin{minipage}[t]{0.51\linewidth}
\vspace{0pt}
\centering
\begin{tabular}{@{}lrrr@{}}
\toprule
\multicolumn{4}{@{}l}{\textup{\textbf{Mercari 100K}}} \\
\multicolumn{4}{@{}l}{\emph{$R^2$\,$\uparrow$ \ ($25{,}000$ test items)}} \\
\midrule
Columns & Cols & Linear & \tabh \\
\midrule
Structured metadata only & 62 & 0.159 & \textbf{0.223} \\
Text only (BGE$\to$PCA-96) & 96 & 0.216 & \textbf{0.434} \\
Fused: metadata $+$ text & 158 & 0.299 & \textbf{0.482} \\
\bottomrule
\end{tabular}
\end{minipage}\\[0.35em]
\parbox{\linewidth}{\raggedright\footnotesize $^{\dagger}$GoEmotions fused adds the five surface counts, including the gold label count.\par}
\end{table}

\paragraph{Many classes and regression.} GoEmotions stresses the label space rather than the text. In Table~\ref{tab:goemotions-main} (left), five surface counts leave every model at the majority class; adding $96$ text columns lifts accuracy by $15$ points, with \tabh{} ahead ($48.11\%$); fusing those columns with the surface counts reaches $51.89\%$. On Mercari 100K log-price regression (right), frozen \tabh{} outperforms ridge on every feature pack, doubling $R^2$ on text alone ($0.434$ vs.\ $0.216$) and reaching $0.482$ when fusing text with metadata.

\paragraph{Text inside a real table.} A major practical application of TFMs is mixed tables where free-text columns sit alongside ordinary structured features---such as an insurance claim with an adjuster's note or a product listing with a description. Standard TFMs typically drop or hash such fields because they have no semantic representation of text. We run one controlled ablation across eight official splits: \tabh{} receives structured columns only, encoded text columns only, or both. LIAR~\cite{wang2017liar} asks for the truthfulness of a political statement (six grades) from its text and metadata (speaker, party, subject, venue). The AutoGluon multimodal suite~\cite{shi2021automm} adds whether a crowdfunding campaign is funded (Kickstarter), whether a job ad is fraudulent (fake jobs), three retail price regressions (American Eagle, JCPenney, Mercari) and a ten-way Airbnb price band.

\begin{figure}[t]
\centering
\includegraphics[width=\textwidth]{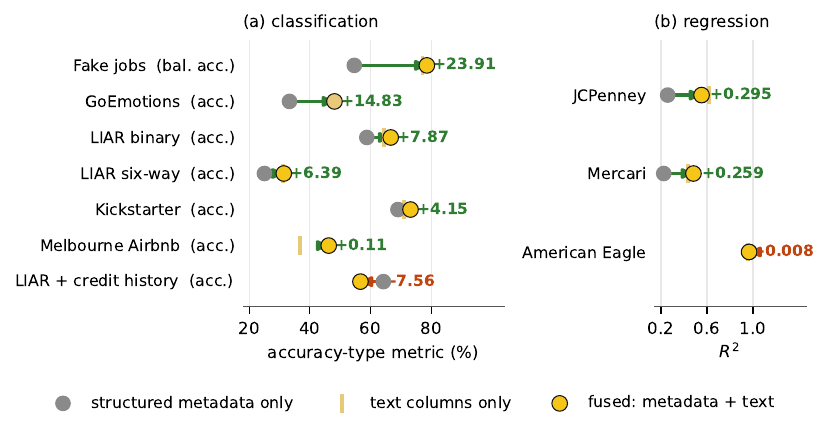}
\caption{\textbf{What is text worth to a TFM that already has a table?} Each row fixes the split, encoder, width ($k{=}96$) and model, and varies only which columns \tabh{} receives: grey is structured metadata alone, the gold tick is text alone, yellow is both. The arrow is the fused-minus-metadata delta, green where text helps and rust where it hurts. Metrics differ across rows; the sign and ordering are the claim. GoEmotions has no structured columns, so its endpoint is text against five surface counts; LIAR $+$ credit history includes a column that leaks the label.}
\label{fig:worth}
\end{figure}

\begin{table}[t]
\centering
\caption{\textbf{Where adding text helps.} \tabh{} on structured columns only, text only, and both. $\Delta$ is fused minus metadata---the quantity a practitioner deciding whether to embed a text field faces. Official test splits.}
\label{tab:when-text}
\newsavebox{\whentextbox}
\sbox{\whentextbox}{%
\small
\setlength{\tabcolsep}{4.5pt}%
\begin{tabular}{@{}llrrrrr@{}}
\toprule
Dataset & Metric & {Test rows} & {Meta.} & {Text} & {Fused} & {$\Delta$} \\
\midrule
\multicolumn{7}{l}{\emph{text pays: the structured columns are ordinary metadata}} \\
Fake job postings & bal.\ acc.\ (\%) & 3{,}182 & 54.64 & 77.08 & \textbf{78.56} & $+23.92$ \\
LIAR binary & accuracy (\%) & 1{,}283 & 58.77 & 64.54 & \textbf{66.64} & $+7.87$ \\
LIAR six-way & accuracy (\%) & 1{,}283 & 25.02 & 31.25 & \textbf{31.41} & $+6.39$ \\
Kickstarter funding & accuracy (\%) & 21{,}626 & 69.02 & 71.10 & \textbf{73.17} & $+4.15$ \\
JCPenney prices & $R^2$ & 2{,}715 & 0.257 & \textbf{0.615} & 0.552 & $+0.295$ \\
Mercari prices$^{\ddagger}$ & $R^2$ & 25{,}000 & 0.223 & 0.434 & \textbf{0.482} & $+0.259$ \\
\midrule
\multicolumn{7}{l}{\emph{text is redundant: the structured columns already determine the target}} \\
Melbourne Airbnb & accuracy (\%) & 4{,}579 & 46.10 & 36.65 & 46.21 & $+0.11$ \\
American Eagle prices & $R^2$ & 5{,}666 & \textbf{0.974} & 0.957 & 0.965 & $-0.009$ \\
\bottomrule
\end{tabular}}%
\usebox{\whentextbox}\par
\vspace{0.35em}
\parbox{\wd\whentextbox}{\raggedright\footnotesize $^{\ddagger}$metadata omits \texttt{brand} and \texttt{cat3} (current GPU memory limit); $\Delta$ is against that reduced table.\par}
\end{table}

\paragraph{Where text boosts performance.} Figure~\ref{fig:worth} and Table~\ref{tab:when-text} split cleanly into two regimes. On six of the eight splits, frozen text embeddings yield substantial gains. Fraud detection gains $23.9$ points of balanced accuracy (frauds caught rise from $13$ to $79$ of $138$). LIAR six-way rises from $25.0\%$ to $31.4\%$, surpassing both the text CNN ($27.0\%$) and the hybrid text-plus-metadata CNN ($27.4\%$) that were \emph{trained} for the task~\cite{wang2017liar}. Kickstarter gains $4.2$ points on a $68\%$ majority baseline, and the two hard price tables gain roughly $0.26$--$0.30$ $R^2$. On the remaining two splits text adds nothing because structured columns already explain the target: American Eagle reaches $R^2=0.974$ from catalog fields, several of which directly encode the price, and Airbnb's listing prose changes only five of $4{,}579$ test predictions. LIAR's shipped credit-history counts show the extreme: they summarise each speaker's truthfulness over the corpus, and with them present text \emph{costs} $7.6$ points. The takeaway is a property of the table itself:

\begin{center}
\setlength{\fboxsep}{7pt}
\fbox{\begin{minipage}{0.93\textwidth}
\textbf{When text pays.} Frozen text repays its columns when the structured fields are \emph{ordinary metadata}---a funding goal, a salary range, a category path. It is redundant when they are already \emph{near-sufficient} for the target, whether because the task is easy or because a column leaks the label.
\end{minipage}}
\end{center}

The check is cheap: fit the metadata-only model first, and embed the text only if that model is far from the achievable ceiling. Conversely, serialising clean numeric columns into prose loses signal, and near-linear semantic similarity tasks leave the TFM no margin over ridge regression (verified on Titanic and STS-B in the supplementary material).

\FloatBarrier

\section{TFMs for vision: Tabular Models can classify images}
\label{sec:pixels}

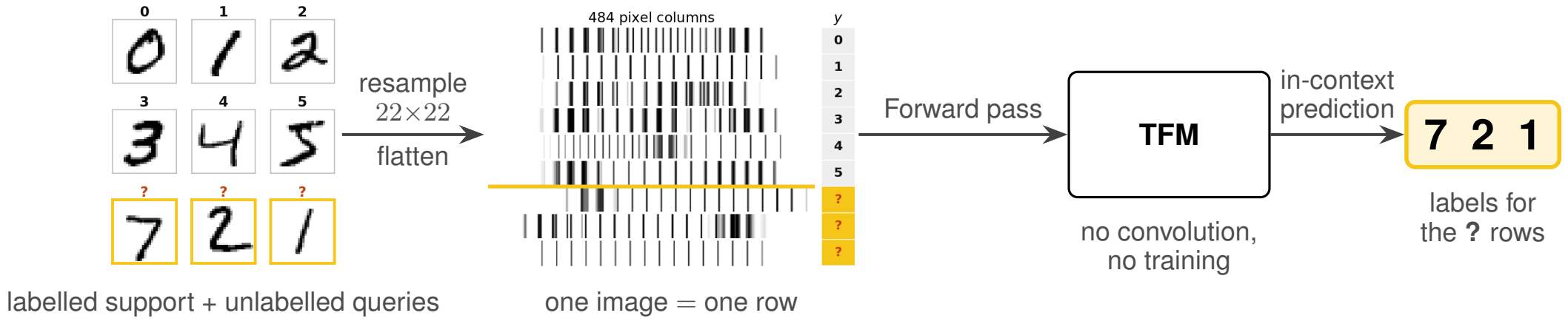
\begin{figure}[!htb]
\centering
\resizebox{\textwidth}{!}{%
\begin{tikzpicture}[
  font=\sffamily\small,
  flow/.style={-{Stealth[length=3mm]}, line width=1pt, black!75},
  lbl/.style={font=\sffamily\footnotesize, align=center, text=black!70},
  tfmbox/.style={draw=black, line width=0.8pt, rounded corners=3pt, fill=white,
               text=black, font=\sffamily\bfseries, align=center,
               minimum height=1.5cm, minimum width=2.4cm},
]
\node[inner sep=0] (grid) at (0,0) {\includegraphics[height=3.2cm]{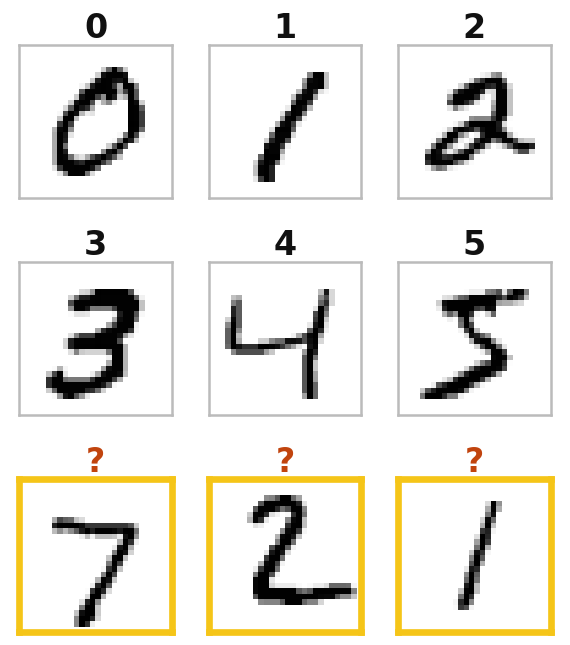}};
\node[inner sep=0, right=1.7cm of grid] (tab) {\includegraphics[height=3.2cm]{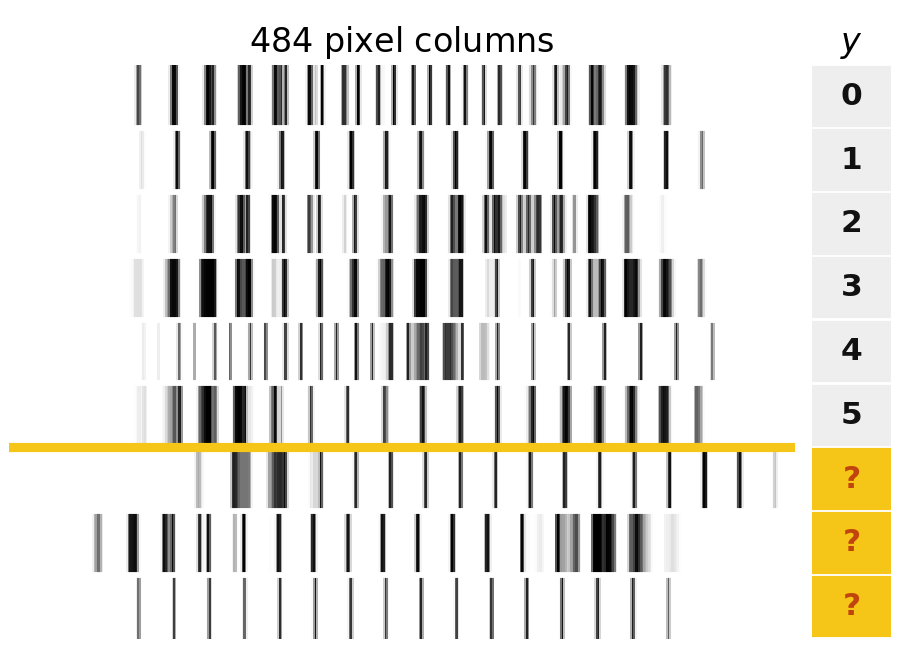}};
\node[tfmbox, right=2.5cm of tab] (tfm) {TFM};
\node[right=1.6cm of tfm, draw=h2oyellow, line width=1.4pt, rounded corners=3pt,
      fill=h2oyellow!18, align=center, inner sep=6pt, font=\sffamily\bfseries\Large] (out)
      {7\;\;2\;\;1};
\draw[flow] (grid) -- node[lbl, above] {resample\\$22{\times}22$} node[lbl, below] {flatten} (tab);
\draw[flow] (tab) -- node[lbl, above] {Forward pass} (tfm);
\draw[flow] (tfm) -- node[lbl, above] {in-context\\prediction} (out);
\node[lbl, below=0.15cm of grid] {labelled support\;+\;unlabelled queries};
\node[lbl, below=0.15cm of tab] {one image $=$ one row};
\node[lbl, below=0.15cm of tfm] {no convolution,\\no training};
\node[lbl, below=0.15cm of out] {labels for\\the \textbf{?} rows};
\end{tikzpicture}}
\caption{\textbf{A TFM reads digits the way it reads any table.} Each MNIST image is resampled to $22\times 22$ and flattened into one row of $484$ intensities. Labelled support images become training rows, test images become rows whose label is \textbf{?}, and a frozen TFM fills in the missing labels by in-context learning. It is never told that the columns form a grid.}
\label{fig:pixel-pipeline}
\end{figure}

A convolutional network is built to know that neighbouring pixels belong together. A TFM is not. It is pretrained on synthetic tables of anonymous columns, and at deployment it standardises each column and averages over column order, so nothing in the model says that column $j$ sits beside column $j{+}1$. We ask whether it can learn that relationship off the shelf: one image becomes one row of intensities, the frozen model predicts the missing label, and no gradient is taken (Figure~\ref{fig:pixel-pipeline}). We demonstrate that frozen TFMs classify MNIST and Fashion-MNIST images directly from pixels.

Each image is resampled to $22\times 22$ and flattened. The support is a balanced draw from the official training split, and the curve uses a sealed $2{,}000$-image test. Chance is $10\%$. On that same support we train logistic regression, a $1$-NN, an MLP on the flattened pixels, and a small CNN on the grid, for $40$ epochs. Nakerst et al.~\cite{nakerst2026nontabular} briefly explored this protocol with TabPFN\,v3 on MNIST and publish the one-shot and $2{,}000$-shot endpoints on all $784$ raw pixels; we repeat it with frozen \tabh{} on the $22\times 22$ canvas and add Fashion-MNIST~\cite{xiao2017fashion}.

\begin{figure}[t]
\centering
\begin{minipage}[c]{0.52\textwidth}
\centering
\begin{tikzpicture}
\begin{axis}[
  width=\linewidth,
  height=5.2cm,
  xlabel={labelled examples per class},
  ylabel={test accuracy (\%)},
  xmin=0.85, xmax=118,
  ymin=0, ymax=100,
  xmode=log,
  log ticks with fixed point,
  xtick={1,5,10,20,50,100},
  xticklabels={1,5,10,20,50,100},
  grid=major,
  grid style={gray!18},
  tick label style={font=\footnotesize},
  label style={font=\footnotesize},
  legend style={font=\scriptsize, at={(0.02,0.98)}, anchor=north west,
                draw=none, fill=white, fill opacity=0.85, text opacity=1, row sep=-1pt},
  legend cell align=left,
]
\addplot[gray!55, densely dashed, thick, forget plot] coordinates {(1,10) (100,10)};
\addplot[color=h2oyellow!80!black, mark=*, thick] coordinates {
  (1,17.20) (5,64.10) (10,78.45) (20,83.40) (50,92.35) (100,93.25)};
\addlegendentry{MNIST \tabh}
\addplot[color=gray!70, mark=square*, mark size=2pt, thick] coordinates {
  (1,43.50) (5,66.65) (10,78.25) (20,82.35) (50,84.65) (100,85.05)};
\addlegendentry{MNIST linear}
\addplot[color=h2oyellow!80!black, mark=triangle*, thick, densely dotted] coordinates {
  (1,9.45) (5,60.05) (10,72.75) (20,75.70) (50,80.00) (100,81.90)};
\addlegendentry{Fashion \tabh}
\addplot[color=gray!70, mark=square*, mark size=2pt, densely dotted, thick] coordinates {
  (1,48.60) (5,66.90) (10,72.00) (20,74.55) (50,74.60) (100,75.80)};
\addlegendentry{Fashion linear}
\addplot[color=trained, mark=otimes*, thick] coordinates {
  (1,37.9) (5,57.4) (10,80.7) (20,89.4) (50,95.5) (100,96.9)};
\addlegendentry{MNIST CNN}
\addplot[color=trained, mark=otimes*, densely dotted, thick] coordinates {
  (1,51.7) (5,61.9) (10,69.2) (20,73.4) (50,79.8) (100,81.0)};
\addlegendentry{Fashion CNN}
\end{axis}
\end{tikzpicture}
\end{minipage}\hfill
\begin{minipage}[c]{0.46\textwidth}
\centering
\footnotesize
\setlength{\tabcolsep}{3.2pt}
\begin{tabular}{@{}llccc@{}}
\toprule
 & & \multicolumn{3}{c}{examples / class} \\
\cmidrule(lr){3-5}
Dataset & Method & $1$ & $100$ & $2{,}000$ \\
\midrule
MNIST & Logistic & \textbf{44.3} & 86.0 & 90.0 \\
 & $1$-NN & 41.8 & 90.4 & 96.2 \\
 & MLP & 29.7 & 89.4 & 97.5 \\
 & CNN, $22{\times}22$ & 37.4 & \textbf{96.9} & \textbf{99.2} \\
 & CNN, $28{\times}28$ & 41.0 & 96.6 & \textbf{99.2} \\
 & \tabh & 16.8 & 95.1 & 98.7 \\
 & TabPFN\,v3 & 22.5 & --- & 98.6 \\
\midrule
Fashion & Logistic & 49.0 & 76.3 & 82.2 \\
 & $1$-NN & 49.1 & 74.8 & 82.5 \\
 & MLP & 36.5 & 78.1 & 87.2 \\
 & CNN, $22{\times}22$ & \textbf{52.0} & 81.9 & 89.8 \\
 & CNN, $28{\times}28$ & 50.7 & 80.1 & \textbf{91.4} \\
 & \tabh & 9.4 & \textbf{83.2} & 89.8 \\
\bottomrule
\end{tabular}
\end{minipage}
\caption{\textbf{A frozen TFM against models trained on the same images.} \textbf{Left:} accuracy on a sealed $2{,}000$-image test as the support grows. Yellow is frozen \tabh, rust is the CNN trained on the same $22\times 22$ canvas, grey is logistic regression on the flattened pixels. Solid is MNIST, dotted is Fashion-MNIST; the dashed line is chance. \textbf{Right:} $1$, $100$ and $2{,}000$ examples per class on $500$ official-test images per class ($5{,}000$ total). The MLP and the $22\times 22$ CNN see the same canvas as \tabh.}
\label{fig:pixels}
\end{figure}

\FloatBarrier

With $100$ labelled images per class, frozen \tabh{} reaches $95.1\%$ on MNIST and $83.2\%$ on Fashion-MNIST (Figure~\ref{fig:pixels}, right). The MLP trained on the same pixels, $127$k parameters and no convolution, reaches $89.4\%$ and $78.1\%$. The TFM is using structure in the rows that a trained flat network does not find.

A CNN, which is built for that structure, is the harder comparison. On MNIST it leads: $96.9\%$ at $100$ shots and $99.2\%$ at $2{,}000$, against $95.1\%$ and $98.7\%$ for \tabh. The $99.2\%$ matches LeNet-5 trained on all $60$k images ($99.05\%$~\cite{lecun1998gradient}). On Fashion-MNIST the TFM matches the CNN at $100$ shots, $83.2\%$ against $81.9\%$, and the CNN on the native $28\times 28$ image leads only at $2{,}000$ shots, $91.4\%$ against $89.8\%$. One example per class is too little context: there a logistic model and a CNN both win. The CNN itself changes by about two points when it is given the native grid instead of the $22\times 22$ canvas.

\begin{figure}[!htb]
\centering
\resizebox{\textwidth}{!}{%
\begin{tikzpicture}[
  font=\sffamily\small,
  flow/.style={-{Stealth[length=3mm]}, line width=1pt, black!75},
  lbl/.style={font=\sffamily\footnotesize, align=center, text=black!70},
  tfmbox/.style={draw=black, line width=0.8pt, rounded corners=3pt, fill=white,
               text=black, font=\sffamily\bfseries, align=center,
               minimum height=1.5cm, minimum width=2.4cm},
]
\definecolor{emb}{HTML}{D0E4F8}
\definecolor{embedge}{HTML}{2E6AAD}
\tikzset{sq/.style={draw=embedge, fill=emb, line width=0.7pt,
                    minimum size=0.36cm, inner sep=0pt, outer sep=0pt}}
\node[inner sep=0] (photos) at (0,0) {\includegraphics[height=4.0cm]{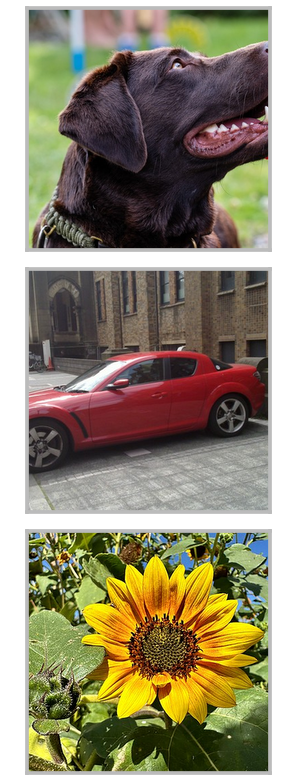}};
\coordinate (clipL) at ([xshift=0.58cm]photos.east);
\coordinate (clipR) at ([xshift=2.15cm]photos.east);
\draw[fill=emb, draw=embedge, line width=0.7pt]
  ([yshift=-1.31cm]clipL) -- ([yshift=1.31cm]clipL)
  -- ([yshift=0.60cm]clipR) -- ([yshift=-0.60cm]clipR) -- cycle;
\path (clipL) -- (clipR) node[midway, font=\sffamily\small, inner sep=0] {CLIP};
\node[minimum width=3.50cm, minimum height=4.0cm, inner sep=0, outer sep=0, anchor=west] (tab) at ([xshift=2.68cm]photos.east) {};
\node[font=\sffamily\footnotesize, inner sep=0, anchor=south] (tabtitle) at ([yshift=1.90cm]tab.center) {$512$ columns};
\foreach \y in {1.073, 0.002, -1.069} {
  \foreach \x in {0.28, 0.70, 1.12, 1.54, 1.96}
    \node[sq] at ([xshift=\x cm, yshift=\y cm]tab.west) {};
  \node[font=\sffamily, inner sep=0] at ([xshift=2.58cm, yshift=\y cm]tab.west) {$\cdots$};
  \node[sq] at ([xshift=3.22cm, yshift=\y cm]tab.west) {};
}
\node[tfmbox, right=2.3cm of tab] (tfm) {TFM};
\node[right=1.6cm of tfm, draw=h2oyellow, line width=1.4pt, rounded corners=3pt,
      fill=h2oyellow!18, align=center, inner sep=5pt, font=\sffamily\bfseries] (out)
      {dog\\[2pt] car\\[2pt] flower};
\draw[flow] (photos.east) -- (clipL);
\draw[flow] (clipR) -- (tab.west);
\draw[flow] (tab) -- node[lbl, above] {Forward pass} (tfm);
\draw[flow] (tfm) -- node[lbl, above] {in-context\\prediction} (out);
\node[lbl, below=0.12cm of photos] {three photographs};
\node[lbl, below=0.12cm of tab] {one image $=$ one row};
\node[lbl, below=0.12cm of tfm] {no training};
\node[lbl, below=0.12cm of out] {predicted\\labels};
\end{tikzpicture}}
\caption{\textbf{High-resolution images take the same path.} CLIP ViT-B/16 maps each photograph to one real $512$-dimensional embedding, so three images are three rows. A frozen TFM reads that table in one forward pass; the labels name the three photographs.}
\label{fig:clip}
\end{figure}
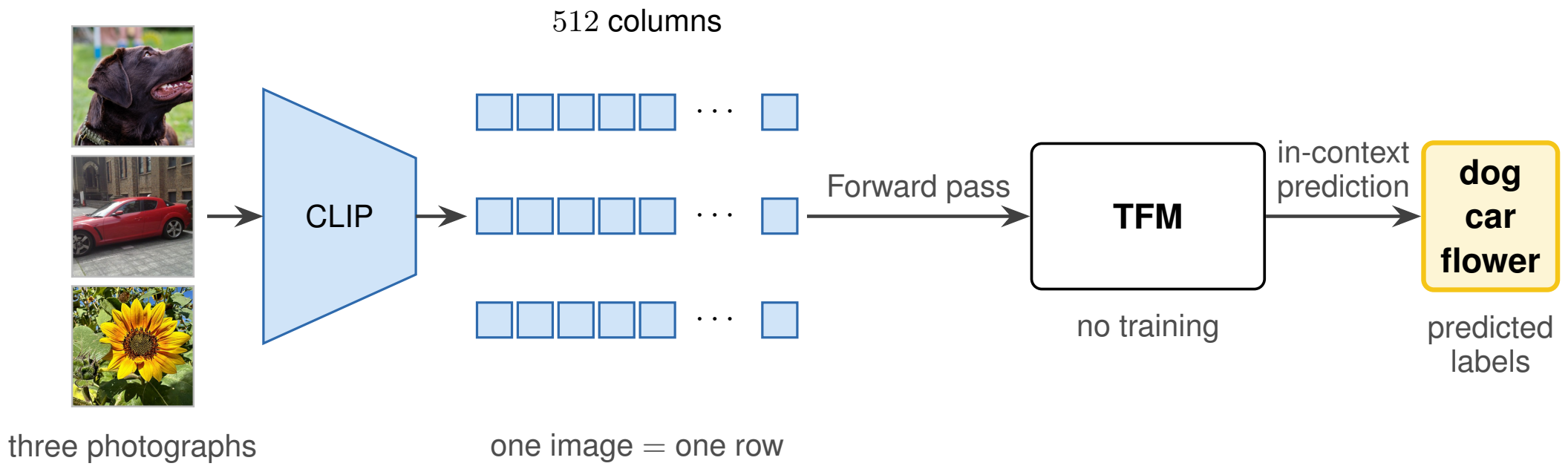

\begin{center}
\setlength{\fboxsep}{7pt}
\fbox{\begin{minipage}{0.93\textwidth}
\textbf{High-resolution images.} A $224\times 224$ colour image is $150{,}528$ numbers, far more columns than a TFM reads in one row on current GPUs. A frozen image encoder removes that constraint (Figure~\ref{fig:clip}). CLIP ViT-B/16~\cite{radford2021clip} maps the whole image to one $512$-dimensional embedding, and those $512$ numbers are ordinary columns. The TFM then classifies the embedding the same way it classifies MNIST, with no projector and no fine-tuning.
\end{minipage}}
\end{center}

\FloatBarrier

\section{TFMs for Motion and Control}
\label{sec:motion}

Motion is already numeric, so no encoder is needed. A window of the recent past---relative positions and velocities, or a robot's state and keypoints---becomes one row, and the target is either the next few waypoints or the next action. Each output coordinate is one in-context regression by frozen \tabh. We test four open-loop settings and one closed loop (Table~\ref{tab:motion}): synthetic curves, where the difficulty is known; LASA human handwriting~\cite{khansari2011lasa} includes non-linear trajectories, which poses a significant problem for TFMs; TrajNet++ pedestrians~\cite{kothari2022trajnetpp}, with sealed test environments; and PushT behaviour cloning~\cite{florence2022ibc,chi2023diffusionpolicy}, with sealed episodes. Every learned baseline sees the same columns as \tabh. The MLP has hidden layers $(256,128)$ and stops on a $10\%$ slice of the training rows. The LSTM is a residual encoder (hidden $128$) whose epoch is chosen by ADE on a $15\%$ slice of the training rows. XGBoost and LightGBM are untuned histogram boosters: $300$ trees of depth $6$, learning rate $0.05$, one regressor per target column. Full splits, published reference points, and per-benchmark figures are in the supplementary material.

\begin{table*}[t]
\centering
\caption{\textbf{Open-loop motion: frozen \tabh{} against controls on identical columns.} Forecasting rows are average displacement error. Synthetic curves and LASA use a five-step horizon. TrajNet++ uses the benchmark horizon of twelve, on the interacting scenes of the sealed holdout ($1{,}042$ scenes). PushT is the $L_2$ error of the predicted action. Lower is better. Naive is constant velocity for forecasting and the mean training action for PushT. XGBoost and LightGBM use $300$ trees of depth $6$. The LSTM cell is empty on PushT because that row is one state, not a history.}
\label{tab:motion}
\small
\setlength{\tabcolsep}{5pt}
\begin{tabular}{lccccccc}
\toprule
Setting & Naive & Ridge & XGBoost & LightGBM & MLP & LSTM & \tabh \\
\midrule
Synthetic curves & 0.350 & 0.353 & 0.326 & 0.333 & 0.313 & 0.348 & \textbf{0.234} \\
LASA handwriting & 0.699 & 0.423 & 0.616 & 0.634 & 0.592 & 0.484 & \textbf{0.403} \\
TrajNet++ interacting & 0.820 & 0.848 & 0.884 & 0.884 & 0.824 & 0.845 & \textbf{0.786} \\
PushT cloning & 131.0 & 21.16 & 18.30 & 18.05 & 19.64 & --- & \textbf{13.58} \\
\bottomrule
\end{tabular}
\end{table*}

\begin{figure*}[t]
\centering
\includegraphics[width=\textwidth]{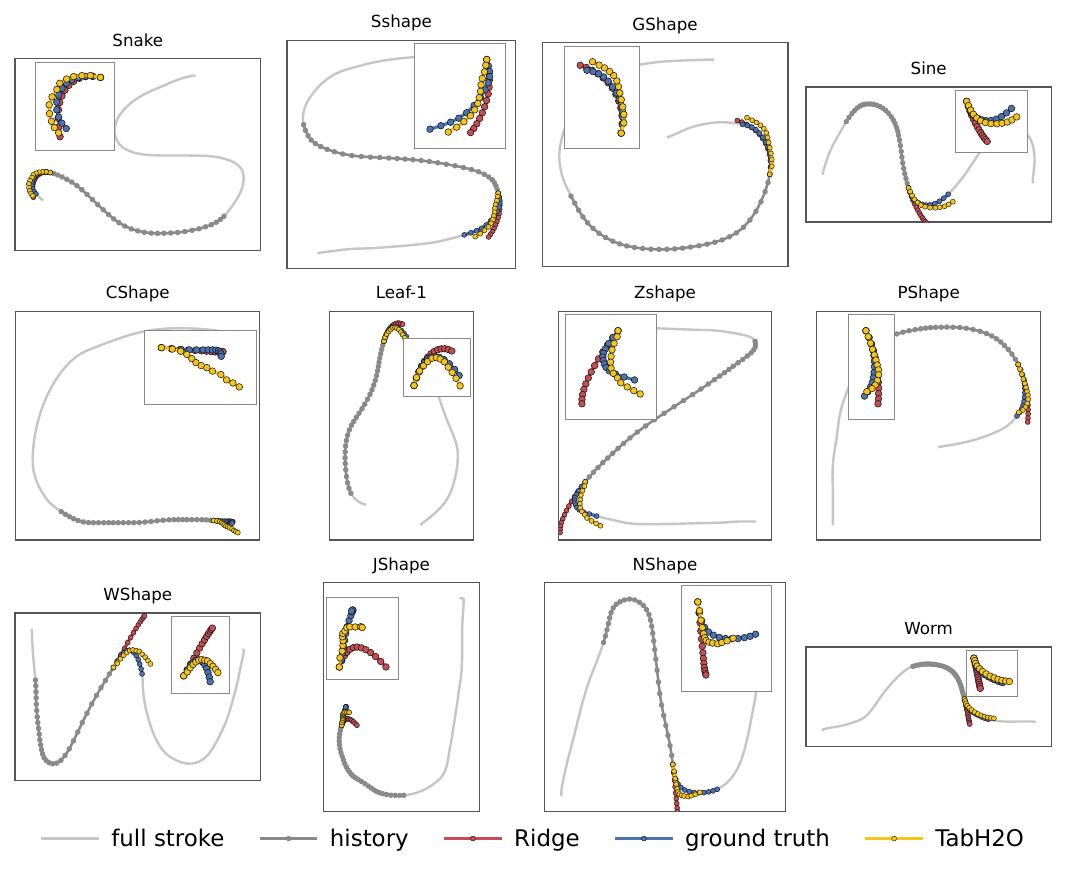}
\caption{\textbf{Frozen \tabh{} forecasts real handwriting (LASA).} Each panel is the held-out demonstration of the named shape, drawn in full so the title is the stroke. Light grey is that demonstration and darker grey is the 16-step history of one window on it. Blue is the next five ground-truth points, yellow the frozen \tabh{} forecast and red ridge; the inset enlarges those five steps. The window is the five-step future with the largest heading change on that stroke.}
\label{fig:lasa}
\end{figure*}

\paragraph{Forecasting and imitation.} Frozen \tabh{} leads every row of Table~\ref{tab:motion}. On synthetic curves the displacement error is $0.234$, a quarter below the MLP ($0.313$). On LASA handwriting it is $0.403$, ahead of ridge ($0.423$); Figure~\ref{fig:lasa} shows the held-out strokes. On PushT the action error is $13.58$, ahead of LightGBM ($18.05$). On TrajNet++ the row is the interacting scenes the benchmark is built to test: constant velocity is $0.820$ and \tabh{} is $0.786$.

\paragraph{Closed-loop control: driving a cave from LiDAR.}\label{sec:cave} Open-loop benchmarks do not test feedback. Following Nakerst's follow-up~\cite{nakerst2026cave}, a 2D car in a procedurally generated cave reads $180$ LiDAR ranges and must steer left, straight or right. A simple follow-the-gap rule provides $500$ labelled rows in one cave; the frozen TFM then drives a \emph{different} cave, receiving one new row per step whose content depends on its own previous action.

\begin{figure}[t]
\centering
\includegraphics[width=\textwidth]{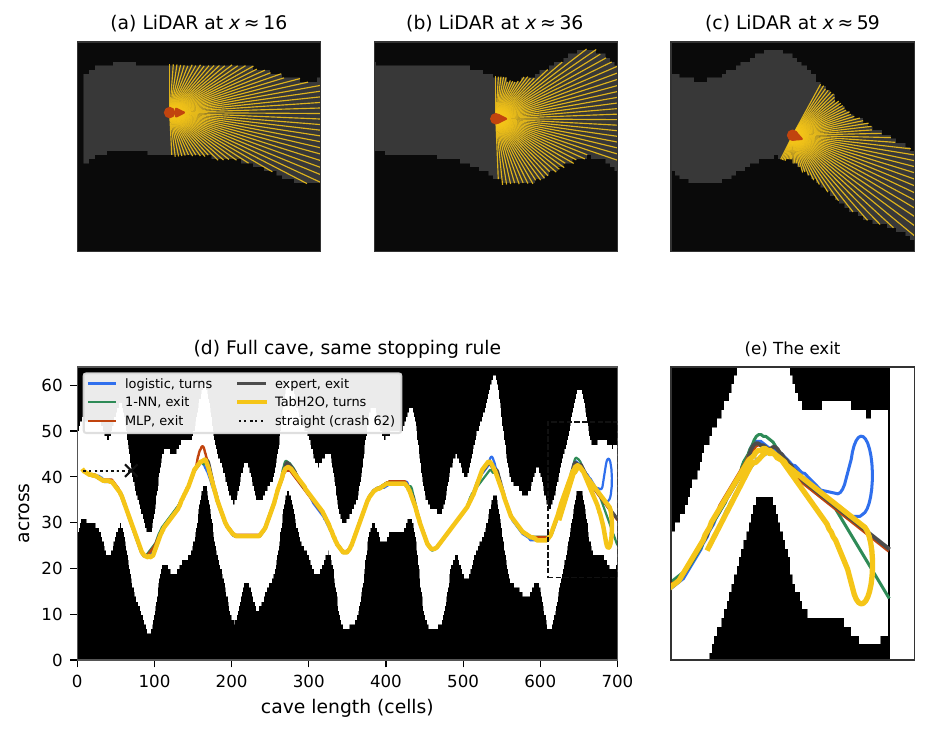}
\caption{\textbf{A frozen TFM driving from LiDAR.} \textbf{(a--c)}~What the controller sees: $180$ ranges (every third ray drawn) at three poses in a held-out cave; dark cells are walls. \textbf{(d)}~Closed-loop traces under one stopping rule: the terminal wall, a crash, or $800$ steps. Grey expert, rust MLP and green $1$-NN reach the end. Yellow \tabh{} and blue logistic regression crest short of it and turn back. The dotted trace crashes. \textbf{(e)}~The exit marked in (d).}
\label{fig:cave}
\end{figure}

\begin{table}[t]
\centering
\caption{\textbf{Cave driving}, $180$ LiDAR columns, three-way steering, one held-out cave. Open loop scores $400$ expert-labelled states. Closed loop rolls every controller until the terminal wall, a crash, or $800$ steps. Max $x$ is the furthest cell reached (the cave ends at $700$). Logistic regression and \tabh{} both crest at $x{=}693$ and are heading back when the budget ends.}
\label{tab:cave}
\small
\setlength{\tabcolsep}{3.4pt}
\begin{tabular}{lccccc}
\toprule
 & \multicolumn{2}{c}{Open loop} & \multicolumn{3}{c}{Closed loop} \\
\cmidrule(lr){2-3}\cmidrule(lr){4-6}
Method & Acc. & Bal. & Result & Steps & max $x$ \\
\midrule
Expert rule & --- & --- & exit & 728 & 699 \\
Always straight & 62.8 & 33.3 & crash & 62 & 70 \\
Logistic regression & 82.8 & 75.4 & turns back & 800 & 693 \\
MLP $(128,64)$ & 87.3 & 81.0 & \textbf{exit} & 727 & 699 \\
$1$-NN & 88.0 & 86.1 & \textbf{exit} & 730 & 699 \\
\textbf{\tabh{} in-context} & \textbf{90.8} & \textbf{86.6} & turns back & 800 & 693 \\
\bottomrule
\end{tabular}
\end{table}

Open loop, \tabh{} recovers the steering rule at $90.8\%$, ahead of $1$-NN ($88.0\%$), an MLP ($87.3\%$) and logistic regression ($82.8\%$) on the same $180$ columns (Table~\ref{tab:cave}). Closed loop asks a harder question: who drives the held-out cave to the end. Always-straight hits a wall at step $62$. The expert reaches the terminal wall at step $728$, and so do the MLP (step $727$) and $1$-NN (step $730$). Logistic regression and \tabh{} both crest at $x{=}693$ and turn around; at step $800$ they are already driving back down the cave (Figure~\ref{fig:cave}). Along the way \tabh{} still matches the expert rule on $83.0\%$ of the states it visits, more often than the MLP ($79.5\%$), logistic regression ($78.8\%$) or $1$-NN ($74.7\%$). Reading the rule and finishing the cave come apart. This is imitation of a short decision rule in simulation. The same frozen checkpoint that grades political statements and reads digits will also \emph{act}, once the observation is a row of numbers.

\section{Related work}
\label{sec:related}

\paragraph{Tabular foundation models.} TabPFN~\cite{hollmann2023tabpfn} introduced prior-data fitted networks~\cite{muller2021pfn} for tabular in-context learning; TabPFN\,v2~\cite{hollmann2025tabpfnv2} added regression; TabICL~\cite{qu2025tabicl} and TabICL\,v2~\cite{qu2026tabiclv2} scale with a column-then-row design; TabDPT~\cite{ma2025tabdpt} pretrains on real tables. \tabh~\cite{pfeiffer2026tabh2o}, TabFM~\cite{google2026tabfm}, EXAONE Tabular~\cite{eo2026exaone} and Causilo~\cite{numsai2026causilo} widen the family. All share per-column standardisation and permutation-averaged inference, the two properties we rely on, so our claims concern the family even though we measure one checkpoint.

\paragraph{Language models and tables.} TabLLM~\cite{hegselmann2023tabllm}, TaBERT~\cite{yin2020tabert} and TabuLa-8B~\cite{gardner2024tabula} serialise tables into text; they excel with very few rows and fall behind TFMs as rows grow. ConTextTab~\cite{spinaci2025contexttab} builds semantics into a new tabular model, which is more expressive and requires pretraining; a \tlm{} touches neither backbone. The AutoGluon multimodal suite~\cite{shi2021automm} supplies our mixed tables; we use it as an ablation of a frozen TFM, not as a comparison with trained multimodal AutoML.

\paragraph{Multimodal alignment and non-tabular TFMs.} LLaVA~\cite{liu2023llava} and Flamingo~\cite{alayrac2022flamingo} established the frozen-encoder-plus-trained-projector recipe; we argue the projector exists only because the downstream model has a learned input manifold. Nakerst et al.~\cite{nakerst2026nontabular,nakerst2026cave} are the closest work and are discussed in Section~\ref{sec:intro}.

\section{Conclusion}

Pretraining on synthetic causal tables teaches tabular foundation models general relational inference that goes well beyond conventional tables. TFMs work directly on structured embeddings in $\Rb^{d}$; thus, if we can encode text, motion, images, or sensor streams into $\Rb^{d}$, a TFM can understand them. Using a single frozen checkpoint (\tabh), we show that this holds across language, vision, motion forecasting, robot cloning, and control, with no trained projector and zero gradient updates. Finally, transfer has clear limits: scarce context, near-linear targets, and distribution shift---that show when in-context inference helps.

\renewcommand{\acksection}{\section*{Acknowledgments}}
\begin{ack}
This work was funded by H2O.ai and is part of the Kaggle Grandmasters Team. Marcos Conde is also with the University of Würzburg. We thank the authors of TabPFN, TabH2O, TabICL, and the maintainers of the public benchmarks used here.
\end{ack}

{
\small
\sloppy
\let\oldthebibliography\thebibliography
\renewcommand{\thebibliography}[1]{\oldthebibliography{#1}\setlength{\itemsep}{2.5pt}}
\bibliographystyle{plain}
\bibliography{references}
}

\clearpage
\appendix

\section{How a TFM sees a table, and why any modality fits}
\label{sec:properties}

The main text evaluates one frozen checkpoint (\tabh) across language, vision, motion, and control. This appendix explains why that cross-domain transfer is structurally well-defined: the exact architecture, pretraining prior, and inference path of \tabh~\cite{pfeiffer2026tabh2o} (Section~\ref{app:sees}), the two invariances that eliminate the need for a trained multimodal projector (Section~\ref{app:invariances}), and the unified column-interface recipe instantiated by every experiment (Section~\ref{sec:method}). No model parameter is updated by gradient descent.

\subsection{Architecture, pretraining prior, and inference path}
\label{app:sees}

All experiments use frozen \tabh\,v1~\cite{pfeiffer2026tabh2o}, a unified $29.2$M-parameter foundation model trained jointly for classification and regression. Building on the three-stage column-then-row design of TabICL~\cite{qu2025tabicl,qu2026tabiclv2}, \tabh{} processes a table of $n$ rows and $m$ features in one forward pass at $O(n^2 + nm^2)$ complexity (Table~\ref{tab:tabh2o-spec}):
\begin{enumerate}[nosep,leftmargin=1.4em]
  \item \textbf{Stage 1 --- Column embedding ($O(nkm)$).} Features are grouped into triples ($G{=}3$) via circular shifts, linearly projected ($d{=}128$), and augmented with target-aware embeddings on training rows. Three induced self-attention blocks (ISAB, $8$ heads, $k{=}128$ learnable inducing vectors) then attend within each column: inducing vectors attend to \emph{training rows only}, and all rows read back from them, preventing query rows from contaminating column statistics.
  \item \textbf{Stage 2 --- Row interaction ($O(nm^2)$).} Three transformer blocks ($8$ heads, rotary position embeddings with base $100{,}000$) attend symmetrically across features within each row. Four learnable CLS tokens collapse each row into a fixed $512$-dimensional vector ($4\times 128$).
  \item \textbf{Stage 3 --- Dataset-wise ICL ($O(n^2)$).} Twelve transformer blocks ($4$ heads, $d_{\mathrm{ICL}}{=}512$) perform block-masked self-attention across rows with keys and values restricted to training rows: training rows attend bidirectionally to one another, while each query row attends to all training rows and never to other queries or itself.
\end{enumerate}
Single-stage training over full-length sequences (without curriculum staging) is stabilised by pre-norm and inter-stage RMSNorm, bounded QASSMax ($\mathrm{scale}_{\mathrm{base}}=c_{\max}\tanh(\mathrm{MLP}(\log n)/c_{\max})$, $c_{\max}{=}10$), learnable residual scaling with stack-input ($x_0$) blending ($\tilde{x}_i=\lambda_{\mathrm{resid}}^{(i)} x_{i-1}+\lambda_{x_0}^{(i)} x_0$), and classification logit soft-capping ($c\tanh(\mathrm{logits}/c)$, $c{=}30$). Dual output heads handle both classification (up to $10$ classes natively, with hierarchical decomposition for higher cardinality such as $28$-way GoEmotions) and regression ($999$ quantiles trained with pinball loss, used for pricing and motion forecasting).

\begin{table}[h]
\centering
\caption{\textbf{Technical summary of the frozen \tabh\,v1 checkpoint~\cite{pfeiffer2026tabh2o} used throughout this paper.} One unified checkpoint handles every classification and regression task with zero gradient updates.}
\label{tab:tabh2o-spec}
\footnotesize
\setlength{\tabcolsep}{5pt}
\begin{tabular}{@{}l p{0.73\textwidth}@{}}
\toprule
Component & Specification \\
\midrule
Backbone & $29.2$M params; $d{=}128$, $d_{\mathrm{ICL}}{=}512$ ($4$ CLS tokens), FFN $2\times$, pre-norm \& inter-stage RMSNorm, bounded QASSMax ($c_{\max}{=}10$), learnable residual scaling with $x_0$ blending \\
Transformer stages & Stage 1: $3$ ISAB ($8$ heads, $128$ inducing, $G{=}3$ circular grouping); Stage 2: $3$ RoPE ($8$ heads, base $100\text{k}$); Stage 3: $12$ block-masked ICL ($4$ heads) \\
Output heads & Unified dual-head: classification (cross-entropy, $\le 10$ native, hierarchical ${>}10$, logit cap $c{=}30$) and $999$-quantile regression (pinball loss) \\
Pretraining prior & Single-stage $100$K steps ($\sim 6.4$M DAG-SCM tables); $2$--$100$ features, $2{,}048$--$12{,}288$ rows, $8$ node function families \\
Noise \& filtering & Up to $3$ noise dims/node ($80\%$ prob.); ExtraTrees filter drops unlearnable tables and trivial ($\mathrm{AUC}{>}0.99$) tables \\
Inference preprocessing & Ordinal categoricals, mean imputation, per-column $z$-scoring, $[-100,100]$ clip, $z$-threshold outlier removal (default $4.0$) \\
Ensemble inference & $N{=}8$ members: Latin-square feature permutations, alternating target flips, random feature sign flips (logit / quantile averaging) \\
\bottomrule
\end{tabular}
\end{table}

\paragraph{Pretraining prior and inference preprocessing.} \tabh{} is pretrained on synthetic DAG structural causal models across $8$ node function families (linear, MLP, quadratic, discretise, tree ensemble, Gaussian process, EM plateau, and product; at most $10$ parents per node; Table~\ref{tab:tabh2o-spec}). Two prior mechanisms matter directly for cross-domain transfer: \emph{noise-aware pretraining} overwrites $\mathrm{Uniform}(0,\min(3,\mathrm{dim}-1))$ output dimensions per node with independent noise ($80\%$ probability), teaching robustness to uninformative trailing coordinates; and \emph{ExtraTrees filtering} removes both unlearnable tables ($\sim 35\%$ classification, $\sim 25\%$ regression) and trivial tables ($\mathrm{AUC}{>}0.99$). At inference, column names are not features, so renaming every column to \texttt{c1, c2, \dots} leaves the network input unchanged, and classification averages logits before softmax while regression averages the $999$ quantiles. Although \tabh{} supports chunked inference and KV caching up to $500{,}000$ rows on one GPU, Pfeiffer et al.~\cite{pfeiffer2026tabh2o} note that multi-modal data such as text, images, or audio sit outside its intended tabular scope.

\paragraph{Four structural limits and transfer boundaries.} The inference path above implies four limits, each visible in our main results:
\begin{enumerate}[nosep,leftmargin=1.4em,label=(L\arabic*)]
  \item \textbf{No column semantics.} Column names are ignored, so raw text is dropped or hashed as a high-cardinality category (a limitation also noted for TabICL\,v2~\cite{qu2026tabiclv2}). Conversely, because column identity carries no constraint, any numerical coordinates can occupy a column slot (Sections~\ref{sec:language}--\ref{sec:motion}).
  \item \textbf{Shape and prior envelope.} Pretraining covers $2$--$100$ features and Stage 2 scales as $O(nm^2)$. That range is also the authors' recommended envelope (about $100$ features; up to roughly $500{,}000$ rows with chunking). Across all benchmarks we evaluate within a uniform GPU memory budget of at most $500$ columns and $100{,}000$ rows; allocating columns to an encoded modality is governed by the width rule in Section~\ref{app:invariances}. The $484$-column pixel tables therefore sit outside the pretraining envelope and inside the evaluation budget.
  \item \textbf{Exchangeability.} Equation~\eqref{eq:pfn} trains for Bayes-optimal prediction under exchangeable context and query rows, and ExtraTrees filtering removes unlearnable tables from the prior. Under covariate shift---as on TweetEval hate (Section~\ref{sec:language}), whose test split was collected separately---\tabh{} commits to non-transferable structure and over-predicts \texttt{hate}. The same exchangeability assumption fails for earlier TFMs on grouped and temporal splits~\cite{jager2026tabpfn35}.
  \item \textbf{Discriminative posterior output.} The heads output class probabilities or $999$ regression quantiles, not free-form text or explanations; every task here is supervised prediction of $y$.
\end{enumerate}
Two task-side boundaries also appear in the main text: \emph{scarce context}, where one labelled image per class leaves too few support rows and logistic regression or a CNN wins (Figure~\ref{fig:pixels}); and \emph{near-linear or already-sufficient structure}, where structured fields already determine the target (American Eagle pricing, Table~\ref{tab:when-text}) or the target is nearly linear in the embedding, leaving \tabh{} little edge over a linear control.

\subsection{Two invariances, admissible interfaces, and interface width}
\label{app:invariances}

A vision-language model requires a trained projector because its language backbone reads a learned token-embedding manifold~\cite{liu2023llava,alayrac2022flamingo}. A TFM has no learned token manifold: after preprocessing, predictions depend strictly on numerical row geometry. Two properties make this exact for \tabh.

\paragraph{P1 (exact positive affine invariance, with ensemble sign symmetry).} For $a\in\Rb^{m}_{>0}$ and $b\in\Rb^{m}$, let $X' = X\,\mathrm{diag}(a) + \mathbf{1}b^{\!\top}$. Per-column $z$-scoring $\sigma(X)_{ij} = (X_{ij}-\mu_j)/s_j$ satisfies $\sigma(X')=\sigma(X)$, since $\mu'_j = a_j\mu_j + b_j$ and $s'_j = a_j s_j$. Both $[-100,100]$ clipping and $z$-threshold outlier removal ($4.0$) act exclusively on $\sigma(X)$, so the forward pass is unchanged: $q_\theta(\cdot\mid X') = q_\theta(\cdot\mid X)$. Units, offsets, and per-column scales never reach the transformer, and negative scalings ($a_j < 0$) are symmetrised in expectation by random feature sign flips across the $N{=}8$ ensemble. TabPFN-3.5~\cite{jager2026tabpfn35} keeps the standardised value, adds learned Fourier features of that value, and adds an in-context ECDF rank invariant to any strictly increasing column transform. It drops quantile and robust rescaling, and still averages eight permutation estimators (four in the fast checkpoint).

\paragraph{P2 (column-order invariance in expectation over the Latin-square ensemble).} In a single forward pass, Stage 1's shared ISAB weights act identically on each column, whereas two mechanisms introduce column-order dependence: repeated circular feature grouping into triples ($G{=}3$) entering Stage 1, and RoPE along the feature axis in Stage 2. \tabh{} restores column-order symmetry at inference by averaging $N{=}8$ ensemble members whose feature orders are rows of a Latin square, placing each feature in $N$ distinct positions across members. The deployed predictor is therefore permutation-invariant in expectation, up to $O(N^{-1/2})$ Monte-Carlo ensemble variance.

\paragraph{Corollary (admissible interfaces and what is ruled out).} By P1 and P2, what \tabh{} reads from a block of columns depends only on the row point cloud modulo per-column affine maps and coordinate permutations---never on column names, units, or positions. Any measurable map $\Phi:\mathcal{T}\to\Rb^{k}$ into a fixed, metrically meaningful coordinate system is therefore admissible without a trained projector, and interfaces related by permuted positive rescalings are equivalent. Raw token IDs fail both tests: coordinate $j$ (``the $j$-th token'') shifts whenever a word is inserted, and token IDs are nominal rather than metric, rendering the affine structure preserved by P1 meaningless. Pooled sentence embeddings qualify; raw tokenisation does not.

\paragraph{Interface width and the spectrum rule.} Admissibility does not fix $k$, and every extra column carries a cost. Let $\Sigma$ be the training covariance of the encoder output with eigenvalues $\lambda_1\ge\lambda_2\ge\cdots$, and define the retained-variance fraction
\begin{equation}
\rho(k) \;=\; \sum\nolimits_{i\le k}\lambda_i \;\Big/\; \sum\nolimits_{i}\lambda_i.
\label{eq:rho}
\end{equation}
Increasing $k$ recovers $1-\rho(k)$ variance, but penalises \tabh{} through sample dilution at fixed context size $n$, quadratic row-stage cost $O(nm^2)$, and shift beyond the pretraining support of $2$--$100$ features (Table~\ref{tab:tabh2o-spec}). Even with noise-aware pretraining, hundreds of low-variance coordinates move the table outside the prior amortised by Equation~\eqref{eq:pfn}:

\begin{quote}
\textbf{Width rule.} Select the smallest $k$ whose retained variance $\rho(k)$ enters $[0.7,0.9]$. Widen if $\rho(k)$ sits well below; if a small $k$ already gives $\rho(k)\approx 1$, do not widen, as the embedding has collapsed onto a few directions.
\end{quote}

This rule fixes $k{=}96$ for single-text BGE-small embeddings in Section~\ref{sec:language} (width sweeps appear in the supplementary material), as when MulTaBench reduces non-native embeddings to $30$ PCA components~\cite{jager2026tabpfn35}.

\subsection{Unified recipe and evaluation protocol}
\label{sec:method}

The corollary yields one gradient-free template. Given structured columns $x\in\Rb^{p}$ and a non-tabular input $u\in\mathcal{M}$, a fixed coordinate map $\phi:\mathcal{M}\to\Rb^{d}$ embeds $u$ into a metrically meaningful space, an unsupervised projection $\pi_k$ sets the width, and the resulting columns are concatenated with $x$:
\begin{equation}
  \boxed{\;\hat{y}_\star \;=\; q_\theta\!\Big(y \,\Big|\, \Phi(x_\star,u_\star),\; \big\{\big(\Phi(x_i,u_i),\, y_i\big)\big\}_{i=1}^{n}\Big),\qquad \Phi \;=\; \mathrm{id}\,\oplus\,\big(\pi_k \circ \phi\big).\;}
\label{eq:tlm}
\end{equation}
Only $\pi_k$ is fitted to the target dataset, using training rows alone and zero labels. Every benchmark in the main text instantiates Equation~\eqref{eq:tlm}, differing solely in $\phi$:
\begin{itemize}[nosep,leftmargin=1.2em]
  \item \textbf{Language} (Section~\ref{sec:language}). $\phi$ is frozen BGE-small~\cite{xiao2024bge} (\texttt{BAAI/bge-small-en-v1.5}, $33$M params): one $\ell_2$-normalised $384$-d vector per text field ($512$ word pieces max), chosen once on UCI development rows over MiniLM-L6/L12~\cite{wang2020minilm} without consulting the holdout. $\pi_k$ is a train-fitted randomised PCA ($k{=}96$); sentence pairs first form $[\,|e_1{-}e_2| \,\|\, e_1{\odot}e_2\,]\in\Rb^{768}$. $28$-way GoEmotions uses \tabh's hierarchical multi-class head; retail pricing uses its $999$-quantile regression head.
  \item \textbf{Vision} (Section~\ref{sec:pixels}). On MNIST and Fashion-MNIST, $\phi$ bilinearly resamples each image to $22\times 22$ intensities ($484$ columns, within the $500$-column budget of (L2), whereas native $28\times 28{=}784$ exceeds it) with $\pi_k{=}\mathrm{id}$. High-resolution photographs substitute a frozen image encoder (CLIP ViT-B/16, $512$-d; Figure~\ref{fig:clip}), a three-image illustration above the $500$-column benchmark budget.
  \item \textbf{Motion} (Section~\ref{sec:motion}). $\phi$ stacks a $16$-step window of relative $(x,y)$ positions and finite-difference velocities ($64$ columns), or a robot state and keypoints ($18$ columns on PushT), with $\pi_k{=}\mathrm{id}$. Each coordinate is one call to the regression head.
  \item \textbf{Control} (LiDAR cave, Section~\ref{sec:motion}). $\phi$ is the raw $180$-ray LiDAR scan with $\pi_k{=}\mathrm{id}$: $500$ expert rows from another cave, and one query per closed-loop step.
\end{itemize}

\paragraph{Evaluation protocol.} Official splits are used wherever available (UCI Sentiment uses a sealed stratified holdout). Both $\pi_k$ and all baselines are fitted strictly on training rows, giving \tabh{} and its controls identical columns, with zero gradient updates to \tabh.

\end{document}